\documentclass[pmlr]{jmlr}

\usepackage{xcolor}

\RequirePackage{graphicx}

 \usepackage{booktabs}
 \usepackage{multirow}
 \usepackage{longtable}

\makeatletter
\def\set@curr@file#1{\def\@curr@file{#1}} 
\makeatother
\usepackage[load-configurations=version-1]{siunitx} 

\graphicspath{{Fig/}{./}}

\jmlrproceedings{PMLR}{Proceedings of Machine Learning Research}
\jmlrvolume{340}
\jmlryear{2026}
\jmlrworkshop{Machine Learning for Healthcare}

\title[AID-MAE for Incomplete EHR]{Learning Representations from Incomplete EHR Data with Dual-Masked Autoencoding}

\author{\Name{Xiao Xiang}
       \Email{xiao0605@mit.edu}\\
       \addr Massachusetts Institute of Technology, École Polytechnique Fédérale de Lausanne
       \AND
       \Name{David Restrepo}
       \Email{davidres@mit.edu}\\
       \addr Massachusetts Institute of Technology
       \AND
       \Name{Hyewon Jeong}
       \Email{hyewonj@mit.edu}\\
       \addr Massachusetts Institute of Technology
       \AND
       \Name{Yugang Jia}
       \Email{yugang@mit.edu}\\
       \addr Massachusetts Institute of Technology
       \AND
       \Name{Leo Anthony Celi}
       \Email{lceli@mit.edu}\\
       \addr Massachusetts Institute of Technology, Harvard University, Beth Israel Deaconess Medical Center}

\begin{document}

\maketitle

\begin{abstract}
Electronic health records (EHR) arrive masked. Clinicians order measurements selectively, and any patient table thus contains only a subset of the values that characterize the underlying physiological state. Prior masked modeling approaches on EHR data either impute the table before learning, represent missingness through a dedicated placeholder signal, or optimize solely for imputation, which limits the representations they learn for downstream clinical tasks and carries every unobserved entry through the encoder. We introduce \textbf{AID-MAE}, an \textbf{A}ugmented-\textbf{I}ntrinsic \textbf{D}ual-Masked
Autoencoder that learns directly from incomplete tables by combining the intrinsic mask the record already carries with an augmented mask that hides a subset of observed values for reconstruction during pretraining. Neither type of masked entry enters the encoder, so attention operates only over what was observed. AID-MAE achieves consistent improvements over strong baselines across multiple clinical tasks on two datasets. Across experiments, we discuss that recovering the missing entries is not a prerequisite for learning and show that the representations learned carry clinical structure without supervision.\end{abstract}

\section{Introduction}

Clinical decisions are routinely made from incomplete records, and clinical signals are coupled by the underlying physiology that generates them. For example, a systemic infection triggers inflammation that lowers blood pressure \citep{jarczak2021sepsis}, clinicians administer vasopressors to sustain perfusion \citep{evans2021surviving}, prolonged support can lead to tissue hypoxia and rising lactate, an early warning of organ failure \citep{jozwiak2022vasopressors}, and heart rate, oxygen saturation, and temperature shift alongside, reflecting the same underlying process \citep{mao2018multicentre}. An ordered measurement therefore carries information about the others that were ordered, and about those that were not, so the observed values can place a patient without the record being complete. This motivates our core question: can we learn directly from this partial information, without completing it first?

Electronic Health Records (EHRs) contain irregularly-sampled time series with diverse, feature-specific missingness \citep{li2021imputation}. Existing methods approach this data in different ways to address the irregular sampling of clinical time series, including a set-based representation that encodes time series as triplets of time, variable, and value, accommodating missingness by construction \citep{setfunctions, tipirneni2022self, pmlr-v252-oufattole24a}. Other methods \citep{labach2023duett,restrepo2025representationlearninglabvalues} transformed irregularly-sampled time series onto a regular grid (tabular format) with missing entries. The intrinsic missingness in the resulting EHR tables is pervasive, heterogeneous and contains information \citep{groenwold2020informative}, and existing deep learning approaches exhibit difficulties in outperforming classical methods \citep{shwartz2022tabular}, such as gradient-boosted decision trees \citep{chen2016xgboost} on many supervised tabular benchmarks \citep{grinsztajn2022tree}. 

Masked autoencoding \citep{he2022masked} in the vision literature learns from a deliberately incomplete view, relying on the redundancy among pixels, where a hidden patch is recoverable from its neighbors. In EHR, a tabular format gives the model the same access to a value's neighbors, aligning features with time. Capturing this cross-temporal and cross-feature context \citep{futoma2017improved, che2018recurrent, labach2023duett} is essential for the models to understand the time series data. These models, however, still resolve the empty cells before encoding, whether by imputation or by an explicit missingness signal \citep{labach2023duett}. Feeding the encoder only the observed entries with masked autoencoding could remove this step, which changes both what the model computes over and what it has to learn from.

A small observation strengthens our motivation to investigate further: pretraining AID-MAE on MICE-imputed \citep{article} PhysioNet Challenge 2012 data \citep{silva2012predicting} degrades performance relative to pretraining on the incomplete table directly (Appendix \ref{app:mice_result}).

We hence make the following contributions towards learning representations directly from incomplete records:

\begin{itemize}
    \item We introduce a dual-mask mechanism for EHR table pre-training. An intrinsic mask marks events that were not observed and an augmented mask hides a random subset of observed values. Reconstruction targets come from the augmented subset only, and neither kind of masked token enters the encoder. AID-MAE does not require imputation or a missingness-specific input, and its attention cost follows approximately the number of observed values rather than the strict size of the grid.

    \item We examine what the dual mask contributes beyond outperforming tree-based and self-supervised baselines on downstream tasks across datasets. Since unobserved features never enter the encoder, AID-MAE transfers to a dataset with a different feature set without modification, while maintaining its stronger performance. We also discuss how design choices tuned for imputation quality do not carry the same interpretation in general-purpose representation learning.
    
    \item We provide qualitative analyses of the learned representations. Patient-level embeddings separate ICU admission types without supervision, and feature-level embeddings organize laboratory measurements into clusters consistent with clinical groupings, suggesting that the model captures physiologically meaningful structure beyond what the training objective explicitly requires.
\end{itemize}

\subsection*{Generalizable Insights about Machine Learning in the Context of Healthcare}

Our work provides evidence for learning representations directly from what was observed, rather than imputing incomplete records into artificial completeness before modeling. Representations learned this way outperform both tree-based and prior self-supervised baselines on every task (Table \ref{tab:results}), and transfer to a second EHR dataset despite substantial differences in feature availabilities and patient populations (Table \ref{tab:transfer}), suggesting that learning from incomplete data produces representations robust to the feature heterogeneity that typically hinders multi-site deployment. We also find that design choices tuned for imputation quality do not carry over. A masking schedule that upweights rarely observed features helps imputation benchmarks but leaves downstream result flat (Table \ref{tab:joint_ablation_hyperparams}).

\section{Related Work}

\subsection{Masked Autoencoding for Incomplete Tabular Data}
Masked autoencoders (MAE) \citep{he2022masked} have been adapted to tabular data \citep{majmundar2022met,du2024remasker,kim2025predict}. ReMasker \citep{du2024remasker} applies MAE for self-supervised imputations on data with simulated missingness. PMAE \citep{kim2025predict} further adds a proportional masking schedule that corrects the imbalance in mask sampling propensity, and improves imputation accuracies. Lab-MAE \citep{restrepo2025representationlearninglabvalues}, extends this line to imputing lab values in EHRs. However, imputation accuracy remains the reported objective in all three, and none evaluates the learned representation on clinical downstream tasks. \citet{xu2025lsm} explored representation learning for wearables time series, where gaps arise from device dropout on a regular lattice rather than from ordering decisions over irregularly sampled clinical events.

\subsection{Self-Supervised Learning for EHR Time Series}
Recent self-supervised approaches for EHR time series explore diverse pretext tasks and data representations. STraTS \citep{tipirneni2022self} uses a forecasting pretext task with the set representation \citep{setfunctions} to learn contextual embeddings under sparsity and irregular sampling. EBCL \citep{pmlr-v252-oufattole24a} instead focuses on temporally local information by contrasting representations before and after clinically significant events. Within the masked modeling approaches, Labrador \citep{pmlr-v259-bellamy25a} showed that masked modeling yields meaningful embeddings, while also highlighting the difficulty of surpassing tree-based methods with self-supervised approaches on clinical benchmarks \citep{chen2016xgboost}. DuETT \citep{labach2023duett} alternates attention across time and events to capture dependencies in a matrix input, fills unobserved entries with zeros and then adds a missingness token to mark them, which competes for attention and computation. In contrast, AID-MAE operates directly on incomplete matrices, explicitly leveraging interactions among available features, efficiently learning contextual representations that are well suited for various downstream tasks.

\section{AID-MAE: Augmented-Intrinsic Dual-Masked Autoencoder}\label{sec:3}

\begin{figure}[!t]
  \centering  \includegraphics[width=\textwidth,height=0.27\textheight,keepaspectratio]{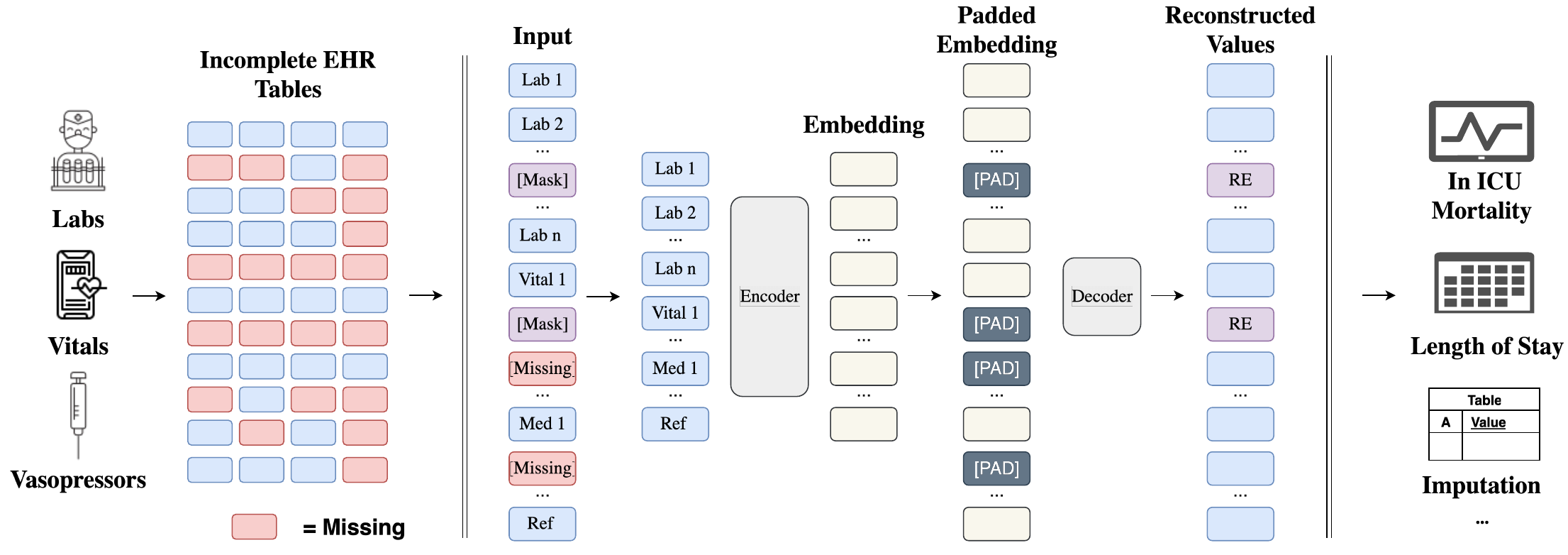}
  \caption{\textbf{AID-MAE (Augmented-Intrinsic Dual-Masked AutoEncoder)} A subset of observed tokens of the original data with inherent missingness \texttt{MISSING} is randomly masked (\texttt{[MASK]}). Each measurement (value and timestamp) is embedded with positional encodings. The encoder processes only unmasked tokens, and the decoder receives the encoded representations along with a learned padding token \texttt{[PAD]} in place of missing or masked entries. Training optimizes a dual loss of reconstructing unmasked values and predicting features under augmented masking, while intrinsically missing entries are excluded from the loss. Code: \url{https://github.com/xiaoxiang0605/AID-MAE}}
  \label{fig:Architecture}
\end{figure}

EHR time series are recorded at irregular time points. Before training, we transform each patient's heterogeneous time-series onto a fixed, ordered grid of length \(L\). The resulting token array
\(
\mathbf{x}\;\triangleq\;(x_1,\dots,x_L)\;\in\;\mathbb{R}^L
\)
is embedded and processed by our model with two distinct masks.

\textbf{Intrinsic Missingness Mask:} Given many slots contain no measurement, we represent this with a binary vector
\begin{align*}
\mathbf{m} \;\triangleq\; (m_1,\dots,m_L)\;\in\;\{0,1\}^L, \quad 
m_i    = 
  \begin{cases}
    1 & \text{if token } i \text{ is recorded},\\[2pt]
    0 & \text{if token } i \text{ is missing}.
  \end{cases}
\end{align*}
On the index set \(\mathcal{I}=\{1,\dots,L\}\), we define recorded set $R$ and missing set $M$ as:\
\[
R := \{\,i\in\mathcal{I}: m_i=1\,\} \quad \ M := \mathcal{I}\setminus R = \{\, i \in \mathcal{I} : m_i = 0\,\}
\] 

\textbf{Augmented Mask:} Given the intrinsic missingness mask \(\mathbf{m}\in\{0,1\}^L\), we sample an augmented mask \(\mathbf{m}'\in\{0,1\}^L\) on the recorded tokens, where \(m'_i=1\) if token \(i\) is kept and \(m'_i=0\) if it is hidden by the augmented mask. We define:
\[
R \setminus A \;=\; \{\, i \in \mathcal{I} : m_i = 1 \ \wedge\ m'_i = 1 \,\},
\quad
A \;=\; \{\, i \in \mathcal{I} : m_i = 1 \ \wedge\ m'_i = 0 \,\}
\]
where \(R \setminus A\) contains tokens that are unmasked and  \(A\) contains tokens hidden by augmented masks.

Building on the masked autoencoder paradigm, our architecture consists of an encoder-decoder Transformer with fixed sinusoidal positional encodings for each feature position \citep{c:22,du2024remasker}. Given per-token embeddings \(\mathbf{z}_i \in \mathbb{R}^d\), only unmasked features \(\{\, \mathbf{z}_i : i \in R \setminus A \,\}\) enter the encoder \(f\), whose layers comprise multi-head self-attention, feed-forward blocks, residual connections, and layer normalization.

Notably, most prior work on masked autoencoding for incomplete tables include unmasked tokens by cropping each array in the batch to the minimum observed input array length \citep{du2024remasker,kim2025predict}, which would remove significantly more measurements for longer sequences, often coming from high acuity patients \citep{agniel2018biases}. In practice, this reduces, for example, a septic patient with hourly blood gases and a full daily laboratory panel to the same length as a stable patient with a single morning draw, discarding precisely the dense monitoring that makes critically ill stays most informative. We instead fix length at the batch maximum \citep{restrepo2025representationlearninglabvalues}, denoted as \(\ell_{\mathrm{keep}}\) and right pad shorter rows with a dedicated token to this length. We then apply a binary attention mask \(\Gamma\in\{0,1\}^{\ell_{\mathrm{keep}}\times \ell_{\mathrm{keep}}}\) so that padded tokens receive zero attention. This preserves equal loss per measurement, prevents pad interactions, and keeps compute at  \(O(\ell_{\mathrm{keep}}^{2})\) per batch.

\textbf{Input Representation:} Each feature (both numerical value and its associated numerical time) is linearly projected to a d-dimensional embedding. We combine the value embedding and time embedding for every measurement. Let
\(
\mathbf{X}\in\mathbb{R}^{L\times d}
\)
denote the ordered token array of length \(L\) with embedding size \(d\), we add positional information to form the initial embedded array before masking:
\(
\mathbf{Z} := \mathbf{X} + \mathbf{P},
\)
where \(\mathbf{P}\in\mathbb{R}^{L\times d}\) denotes the positional encoding. The token indices define a fixed order for features columns. 

\textbf{Reconstruction:} 
Denote the encoder outputs by \(\mathbf{h}_i \in \mathbb{R}^d\) and stack them as
\(
\mathbf{H} = \bigl( \mathbf{h}_i \bigr)_{i \in R \setminus A} 
= f\bigl(\{ \mathbf{z}_i : i \in R \setminus A \}\bigr) 
\;\in\; \mathbb{R}^{\ell_{\mathrm{keep}} \times d},
\) \
where \(\mathbf{z}_i \in \mathbb{R}^d\) is the embedding of token \(i\), \(\ell_{\mathrm{keep}}\) is the maximum length of the fed input arrays in a batch,  and \(\mathbf{h}_i\) denotes its encoded representation. We pad a shared learnable mask token \(\mathbf{m}_{\mathrm{token}}\in\mathbb{R}^d\) at all masked positions, i.e.\ at both intrinsic missing positions \(M = \mathcal{I}\setminus R\) and positions with augmented mask \(A \subseteq R\). We once again add positional encoding:
\[
\mathbf{z}^{\mathrm{dec}}_i \;=\;
\begin{cases}
\mathbf{h}_i, & i\in R\setminus A,\\
\mathbf{m}_{\mathrm{token}}, & i\in A \cup M
\end{cases} \quad
\mathbf{Z}_{\mathrm{dec}} \;=\; \bigl(\mathbf{z}^{\mathrm{dec}}_1,\dots,\mathbf{z}^{\mathrm{dec}}_L\bigr)^\top + \mathbf{P}
\]

The decoder \(g\) predicts the original values at positions with both augmented masks and intrinsic masks, as in Figure \ref{fig:Architecture}.

\textbf{Training Objective:} The model is pretrained by optimizing dual reconstruction loss, which reconstructs features in both set of unmasked tokens \(R \setminus A\) and augmented masks \(A\), while ignoring intrinsic missing entries \citep{du2024remasker}. The reconstruction loss for a training example is defined as:
\begin{align*}
\mathcal{L}
&= \tfrac{1}{|R(\mathbf{m}) \setminus A(\mathbf{m}, \mathbf{m}')|}
   \sum_{i\in R(\mathbf{m}) \setminus A(\mathbf{m}, \mathbf{m}')}(\hat{x}_i - x_i)^2 
+ \tfrac{1}{|A(\mathbf{m}, \mathbf{m}')|} 
   \sum_{i\in A(\mathbf{m}, \mathbf{m}')}(\hat{x}_i - x_i)^2
\end{align*}
The separation of intrinsic and augmented masks is motivated by a key asymmetry. Augmented-masked entries have known ground-truth values and provide valid reconstruction targets, whereas intrinsically missing entries do not. Including intrinsically missing positions in the encoder input or the reconstruction loss would require imputing targets that no clinician chose to measure, introducing potentially unreliable supervision into the training signal.

\section{Experiments}\label{sec:4}

\subsection{Data and Tasks}

\paragraph{Data} We have selected and preprocessed the most frequently sampled 50 laboratory values, five vital signs (heart rate, respiratory rate, systolic  blood pressure, diastolic blood pressure, and temperature), oxygen saturation, and five vasopressors (norepinephrine, epinephrine, vasopressin, dopamine, and phenylephrine) from MIMIC-IV ICU \citep{johnson_mimic-iv_2023}. For the second dataset, PhysioNet Challenge 2012 \citep{silva2012predicting}, we included 23 lab values and 4 vital signs that are also included in our curated MIMIC-IV dataset. Each event contains an item ID, a timestamp, and a numerical value. We first cap each numeric feature at the 5th and 95th percentiles (winsorization) to reduce extreme outliers \citep{wilcox2011introduction}. We then apply a per-feature min--max normalization. If the values are not recorded, those values are represented as a missing entry. We provide a full list of included features in Table \ref{tab:all_features}.

In particular, we applied signal-specific preprocessing for each type of events. For each input array:
\begin{itemize}

    \item{\textbf{Laboratory values}} were selected with a daily frequency. We include a reference value for each lab, which corresponds to the most recent corresponding lab result prior to the day. This aims to align with real-life clinical setting and provide a longitudinal baseline.

    \item{\textbf{Vital signs}} were included with finer granularity of 1 hour. Each vital sign constitutes 24 data points in an input array. 

    \item{\textbf{Vasopressors}} were carried forward for each of the recorded dosages from its start time till its end time. Vasopressor data are not available in the PhysioNet Challenge 2012 dataset, inducing a feature-level 
    distribution shift that allows us to assess robustness and cross-dataset generalization.
\end{itemize}

We include all ICU patient stays, discarding input arrays with very few measurements or without any laboratory measurement. The final datasets comprise 92{,}938 stays spanning 412{,}365 patient-days for MIMIC-IV, and 11{,}987 stays with 23{,}682 patient-days in PhysioNet Challenge 2012. More detailed information on the data pre-processing can be found in Appendix \ref{apd:second}.

\paragraph{Task Definitions} For downstream evaluation, we construct one sample per patient by restricting inputs to the first 24 hours of the ICU admission, following common benchmarks \citep{purushotham2018benchmarking,johnson2018real, hempel2023prediction, yeh2024early}. This design reflects clinical practice, where physicians perform early assessments within hours of admission and subsequently refine decisions as new measurements become available \citep{rivers2001early}.

We evaluate model performance on the following downstream clinical prediction tasks:

\begin{itemize}
\item \textbf{Mortality}: A binary indicator of death during the admission. On MIMIC-IV we use death before ICU discharge; on PhysioNet Challenge 2012 we use the in-hospital outcome provided with the dataset.
\item \textbf{Length of Stay (LOS)}: A binary indicator denoting whether ICU length of stay is strictly less than 72 hours from admission.
\item \textbf{Acute Kidney Injury (AKI)}: Defined using serum creatinine measurements according to the criteria proposed by \citet{10.1159/000339789}. Patients without any creatinine measurement were excluded from AKI evaluation.
\end{itemize}

Mortality and LOS are evaluated on MIMIC-IV, while mortality and AKI are evaluated on the PhysioNet Challenge 2012 dataset. Task label distributions are reported in Table~\ref{tab:cohort-tasks}:

\begin{table}[ht]
  \centering
  \caption{Downstream tasks and cohort sizes for MIMIC-IV and PhysioNet Challenge 2012.}
  \label{tab:cohort-tasks}
  \begin{minipage}{0.48\textwidth}
    \centering
    \textbf{MIMIC-IV}
    \vspace{0.3em}
    
    \begin{tabular}{lcc}
      \toprule
      \textbf{Task} & \textbf{\# Stays} & \textbf{Prevalence} \\
      \midrule
      Mortality   & 92{,}938 & 7.6\% \\
      LOS \(<72\)h & 92{,}938 & 66.1\% \\
      \bottomrule
    \end{tabular}
  \end{minipage}
  \hfill
  \begin{minipage}{0.48\textwidth}
    \centering
    \textbf{PhysioNet Challenge 2012}
    \vspace{0.3em}
    
    \begin{tabular}{lcc}
      \toprule
      \textbf{Task} & \textbf{\# Stays} & \textbf{Prevalence} \\
      \midrule
      Mortality & 11{,}981 & 14.3\% \\
      AKI       & 11{,}811 & 29.3\% \\
      \bottomrule
    \end{tabular}
  \end{minipage}
\end{table}

\subsection{Results} \label{sec: results}

\paragraph{AID-MAE Outperforms All Baselines on
Fine-Tuning Tasks}

Table \ref{tab:results} shows that our model outperforms a range of baselines consistently on all tasks evaluated. This includes XGBoost \citep{chen2016xgboost}, as well as masked modeling method \citep{labach2023duett}. We note that the supervised transformer is trained with the same architecture, with weights initialized at random and updated only through task-specific supervision. This demonstrates that pre-training is an essential component in the superior performance over purely supervised training. Notably, AID-MAE achieves these results with under one million trainable parameters compared to 5.2 million in DuETT, suggesting the gains are attributable to the architectural design rather than increased model capacity. We report fine-tuning details of our model and the baseline models in Appendices \ref{app:architecture} and \ref{app:baseline}, and the complete result in Table \ref{tab:mimic_downstream_results}.

\begin{table*}[t]
  \centering
  \setlength{\tabcolsep}{6pt}
    \caption{Fine-tuning results on MIMIC-IV and PhysioNet Challenge 2012. 
  We report AUROC (mean $\pm$ SD over 5 seeds). Best scores per column are in \textbf{bold}. AID-MAE outperforms all baselines.}
  \begin{tabular}{@{}lcccc@{}}
    \toprule
    Model & \multicolumn{2}{c}{MIMIC-IV} & \multicolumn{2}{c}{PhysioNet Challenge 2012} \\
    \cmidrule(lr){2-3} \cmidrule(lr){4-5}
          & Mortality & LOS & Mortality & AKI \\
    \midrule
    Logistic Regression     & 80.7 $\pm$ 0.0  & 68.8 $\pm$ 0.0  
                            & 72.7 $\pm$ 0.0  & 74.8 $\pm$ 0.0 \\
    Supervised Transformer  & 83.9 $\pm$ 0.5  & 72.3 $\pm$ 0.5  
                            & 76.7 $\pm$ 1.4  & 76.1 $\pm$ 0.9 \\
    XGBoost                 & 86.7 $\pm$ 0.0  & 76.9 $\pm$ 0.0  
                            & 76.9 $\pm$ 0.0  & 77.1 $\pm$ 0.0 \\
    DuETT                   & 86.4 $\pm$ 0.2  & 77.2 $\pm$ 0.0  
                            & 77.7 $\pm$ 0.4  & 76.7 $\pm$ 0.2 \\
    \textbf{AID-MAE (ours)} & \textbf{87.7 $\pm$ 0.1} & \textbf{77.6 $\pm$ 0.1} 
                            & \textbf{78.2 $\pm$ 0.3} & \textbf{77.3 $\pm$ 0.1} \\
    \bottomrule
  \end{tabular}
  \label{tab:results}
\end{table*}

\begin{figure}[!h]
\centering
\includegraphics[width=0.95\textwidth]{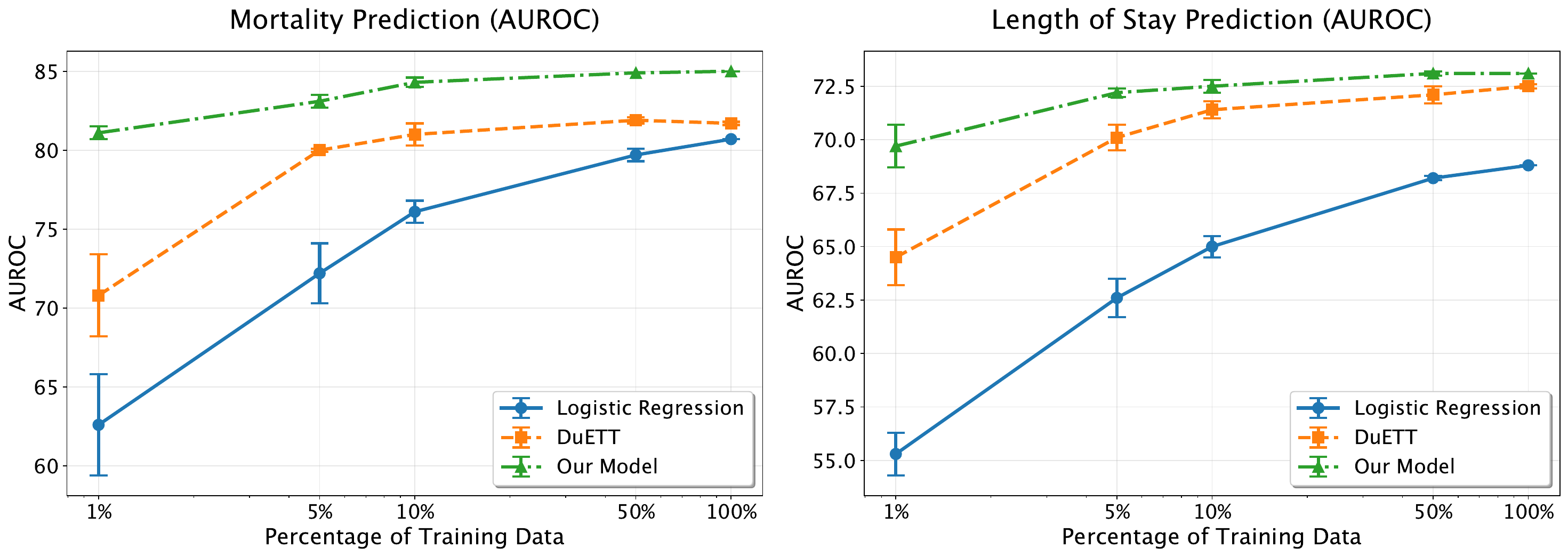}

\caption{Linear probing results for mortality prediction and length of stay prediction tasks. We compare our model against Logistic Regression with raw data (median imputed) and DuETT with frozen encoder, across different training data percentages (1\%, 5\%, 10\%, 50\%, 100\%). Error bars represent standard deviation across 5 seeds.}
\label{fig:linear_probing_results}
\end{figure}

\paragraph{Linear Probing Shows Pretrained Embeddings Transfer Strongly Across Data
Regimes}
Figure \ref{fig:linear_probing_results} shows that our model achieve superior results compared to the DuETT baseline \citep{labach2023duett} in all data availabilities. We additionally include logistic regression with median imputation on raw data as a reference. The performance gap is pronounced in low-data regimes, suggesting our pretrained embeddings encode informative inductive biases, especially when labeled samples are scarce. The full linear probing results are shown in Figure \ref{fig:linear_probing_full}. We additionally note that linear probing asks how much useful information is encoded in the model's learned patient representations, without any further training.

\paragraph{Learned Laboratory Feature Embeddings Exhibit Clinically Coherent Organization}

We analyze whether the learned encoder organizes laboratory measurements into coherent regions of representation space. 
For each laboratory feature, we extract its embedding from a pretrained AID-MAE model on MIMIC-IV and project a random subset of \(N=100{,}000\) embeddings into two dimensions using UMAP \citep{mcinnes2018umap} with a Euclidean metric.

\begin{figure}[!ht]
    \centering
    \includegraphics[width=0.55\linewidth, trim={0 0 225 50}, clip]{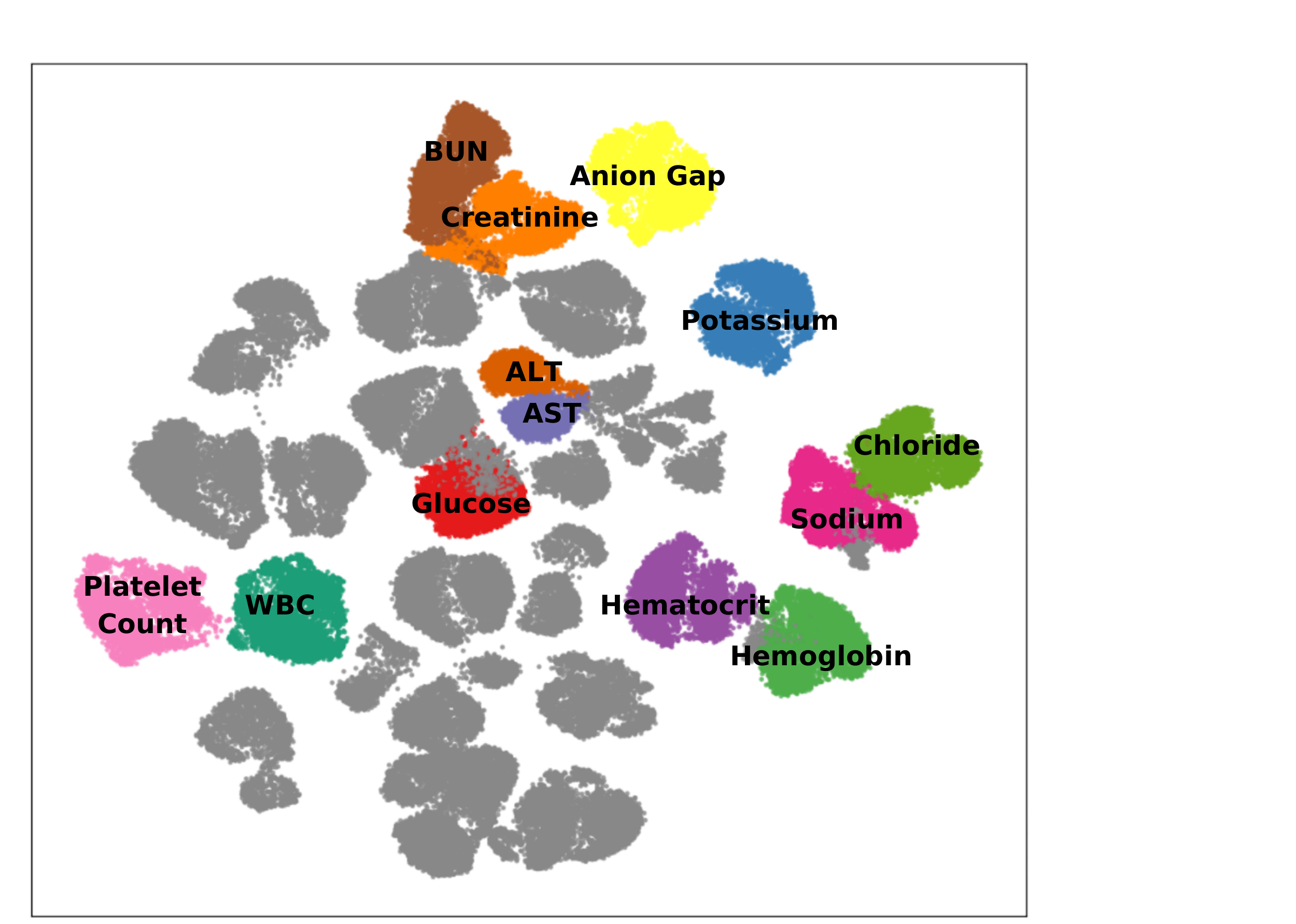}
    \caption{UMAP of feature embeddings. UMAP projection of \(N=100{,}000\) randomly sampled 64-D embeddings for 50 lab features. Colors denote 13 highlighted lab types. Each point corresponds to one measurement embedding. We denote two important patterns: Bottom-left: neighboring pair of pink (Platelet Count) and green islands (WBC); Bottom-right: neighboring pair of purple (Hemoglobin) and green islands (Hematocrit). The geometrical neighboring is consistent with their clinical coupling.}
    \label{fig:umap_features}
\end{figure}

As shown in Figure~\ref{fig:umap_features}, the embedding space exhibits clear, clinically meaningful structure. 
Related laboratory tests form distinct and neighboring clusters, including Platelet Count and White Blood Cell count (both hematology measures), as well as Hemoglobin and Hematocrit, which are tightly coupled in clinical practice. 
Importantly, UMAP is label-agnostic and lab identities are not provided to the model, indicating that these groupings emerge from the learned geometry rather than supervision.

\paragraph{Learned Embeddings Stratify ICU Admissions Into Latent Subtypes}

Beyond linear probing, we explore the extent to which the learned embedding space naturally stratifies patients by physiological state. We assessed this by extracting the CLS embeddings, an input array level summary \citep{devlin2019bertpretrainingdeepbidirectional}, for medical intensive care unit (MICU) and the cardiothoracic vascular intensive care unit (CVICU) admissions. The CLS embeddings successfully differentiate stays admitted to the MICU and CVICU into distinct latent subgroups. This is followed by applying k-means clustering to the original embedding space. 

\begin{figure}[!t]
  \centering
  \includegraphics[width=0.5\linewidth]{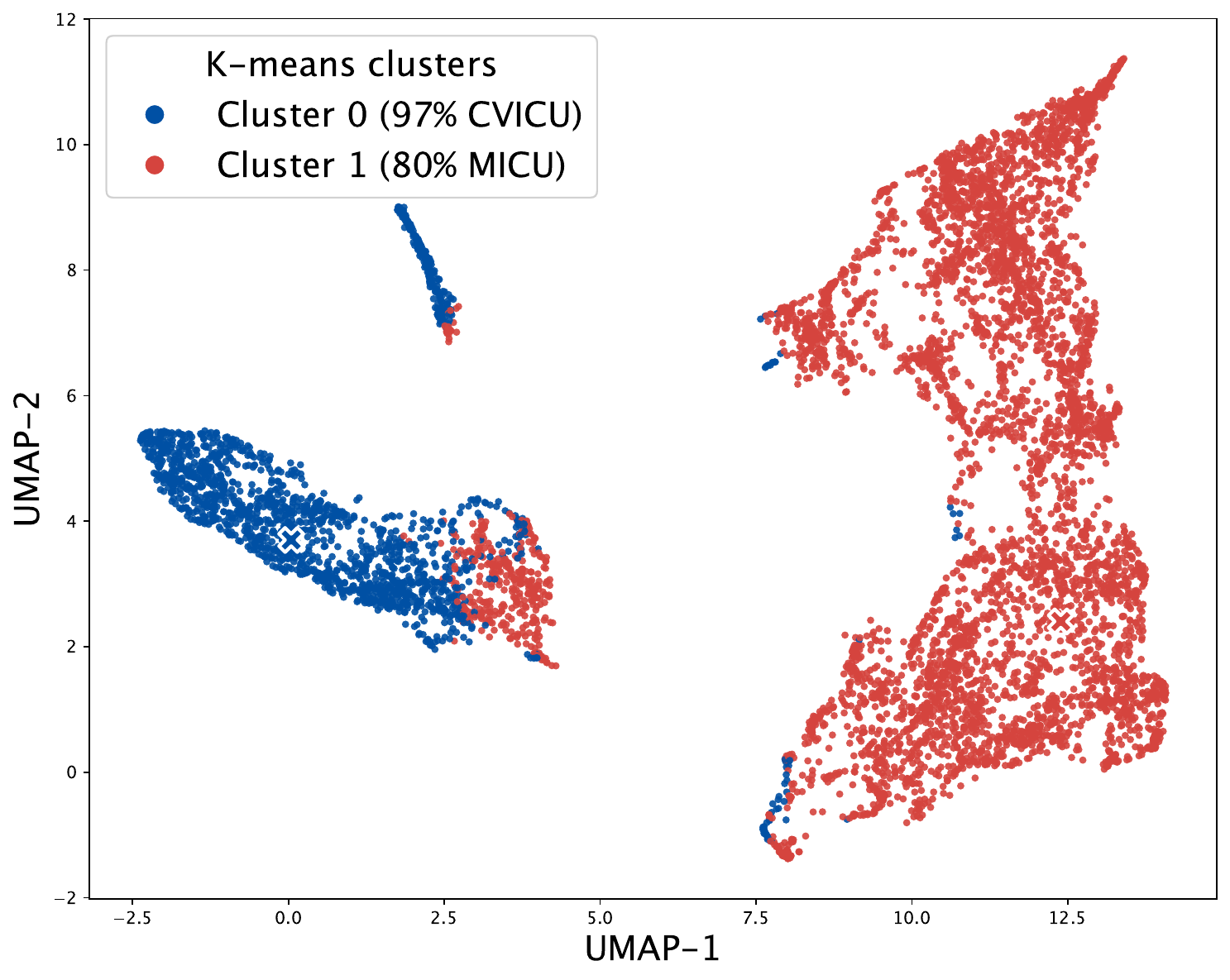}
  \caption{UMAP visualization of first-day CLS embeddings for initial MICU and CVICU admissions. Colors represent clusters from K-means (\(k=2\)) applied in the embedding space.}
  \label{fig:umap_kmeans_cvicu_micu}
\end{figure}

Figure \ref{fig:umap_kmeans_cvicu_micu} reveals two well-separated clusters in the UMAP space \citep{mcinnes2018umap}: Cluster~0 contains \(97\%\) CVICU admissions and Cluster~1 contains \(80\%\) MICU admissions. The high homogeneity of each cluster with respect to ICU type indicates that the learned representations capture clinically meaningful structure reflecting unit-specific physiology and care contexts.

We note that, to determine the optimal number of clusters, we conducted a silhouette analysis over k=2 to 10 clusters. The analysis identified k=2 as optimal, with a silhouette score of 0.42, which indicates moderate but meaningful separation between patient subtypes.

\section{Towards Generalizable EHR Representations}

\subsection{Cross-Dataset Considerations}

Different EHR datasets exhibit significant distribution shifts in patient populations, institutional care practices, and measurement availability \citep{burger2024foundationmodelscriticalcare}. A model pretrained on one dataset will encounter a different set of recorded variables, different missingness rates, and different clinical protocols on another. Understanding how pretrained representations behave under these shifts is important for assessing their practical utility.

One consideration is how the model handles features present during pretraining but absent at evaluation time. DuETT \citep{labach2023duett}, for example, requires absent features to be carried through the encoder as explicit missing tokens to preserve positional compatibility, introducing unnecessary computation. AID-MAE sidesteps this by excluding unobserved features from the encoder, allowing it to operate on whatever subset of features is available without architectural modification.
\\ \\
\textbf{Cross-Dataset Transfer:} To evaluate this, we transfer MIMIC-IV \citep{johnson_mimic-iv_2023} pretrained weights to PhysioNet Challenge 2012 \citep{silva2012predicting}, a dataset with substantially fewer features, no vasopressor data, and a different patient population.
 As shown in Table \ref{tab:transfer}, frozen AID-MAE representations outperform DuETT by 2.2 and 3.5 AUROC points on mortality and AKI under linear probing, suggesting that the pretrained representations capture clinical structure that generalizes beyond the pretraining dataset. Under fine-tuning, AID-MAE maintains consistent gains across both tasks.
 
\begin{table*}[!ht]
\centering
\setlength{\tabcolsep}{6pt}
\caption{Transfer learning performance from MIMIC-IV pre-trained weights to PhysioNet Challenge 2012.
We report AUROC for mortality and acute kidney injury (AKI) prediction under linear probing and full fine-tuning.}
\label{tab:transfer}

\begin{tabular}{llcc}
\toprule
\textbf{Training Regime} & \textbf{Method} & \textbf{Mortality} & \textbf{AKI} \\
\midrule
\multirow{3}{*}{Linear Probing}
& Random Weights          & 49.8 & 52.7 \\
& DuETT                   & 70.9 & 66.2 \\
& \textbf{AID-MAE (ours)} & \textbf{73.1} & \textbf{69.7} \\
\midrule
\multirow{3}{*}{Fine-Tuning}
& Random Weights          & 76.7 $\pm$ 1.4 & 76.1 $\pm$ 0.9 \\
& DuETT                   & 77.8 $\pm$ 0.3 & 76.4 $\pm$ 0.1 \\
& \textbf{AID-MAE (ours)} & \textbf{78.6 $\pm$ 0.2} & \textbf{77.0 $\pm$ 0.2} \\
\bottomrule
\end{tabular}

\end{table*}

\paragraph{Logit-Based Proportional Reweighting}
Driven by differences in ordering practice, missingness varies substantially within a single dataset, and across healthcare centers. Among features in EHRs, some variables are nearly always observed, while others are scarcely recorded \citep{li2021imputation}. In our data, for instance, serum sodium is missing around \(7\%\) of patient--day entries, whereas arterial oxygen saturation is absent in over \(88\%\) of cases. We thus were motivated to test a proportional augmented masking scheme explicitly designed to counterbalance the heterogeneous missingness within a single dataset and potential distribution shifts across datasets. A simple reweighting \(r_j= \pm \ p_{\mathrm{miss},j}\), given how much a feature is missing per batch, may collapse sampling to one regime (dense vs. sparse) \citep{cui2019class,li2019addressing}. We therefore map missing frequencies into weights by using a logit transform, a schedule that has the desired convex-concave curvature \citep{kim2025predict}:

\[
w_j =
\begin{cases}
a\,\log\!\dfrac{p_{\mathrm{miss},j}}{1 - p_{\mathrm{miss},j}} + b,
& \text{if } p_{\mathrm{miss},j} < 1, \\[6pt]
0,
& \text{if } p_{\mathrm{miss},j} = 1.
\end{cases}
\]

where the parameter $a$ determines the resampling direction and offset $b$ serves as a regularizer to avoid 
extreme weights when $p_{\mathrm{miss},j}$ approaches zero. In light of these considerations, we examine whether this correction, developed for imputation, carries over to representation learning in the following ablation studies.

\subsection{Ablation Studies on Masking and Input Design}

\paragraph{Masking Ratio and Reweighting}

Table~\ref{tab:joint_ablation_hyperparams} reports results across masking ratios and reweighting hyperparameters. Downstream AUROC varies under one point over the whole grid, and neither sign of $a$ is consistently better than random masking. We therefore use random masking at 25\% ($a=0$, $b=0.25$) for all reported results.

\begin{table*}[t]
\centering
\small
\setlength{\tabcolsep}{6pt}
\renewcommand{\arraystretch}{1.05}
\caption{Ablation study on proportional masking hyperparameters. Performance is reported as mean $\pm$ standard deviation across 5 random seeds. Downstream performance is largely flat across the grid}
\label{tab:joint_ablation_hyperparams}
\begin{tabular}{cccccc}
\toprule
\multirow{2}{*}{\textbf{$a$}} & \multirow{2}{*}{\textbf{$b$}} &
\multicolumn{2}{c}{\textbf{Mortality}} &
\multicolumn{2}{c}{\textbf{Length of Stay}} \\
\cmidrule(lr){3-4}\cmidrule(lr){5-6}
& & \textbf{AUROC} & \textbf{AUPRC} & \textbf{AUROC} & \textbf{AUPRC} \\
\midrule

\multicolumn{6}{l}{\textbf{Random masking ($a=0$)}} \\
\midrule
0 & 0.125 & 87.4 $\pm$ 0.1 & 49.4 $\pm$ 0.2 & 76.8 $\pm$ 0.2 & 62.0 $\pm$ 0.1 \\
0 & 0.25  & 87.7 $\pm$ 0.1 & 49.8 $\pm$ 0.1 & 77.6 $\pm$ 0.1 & 63.1 $\pm$ 0.1 \\
0 & 0.5   & 87.3 $\pm$ 0.1 & 49.0 $\pm$ 0.1 & 77.2 $\pm$ 0.2 & 62.0 $\pm$ 0.3 \\
0 & 0.75  & 87.5 $\pm$ 0.1 & 49.8 $\pm$ 0.2 & 76.8 $\pm$ 0.2 & 61.9 $\pm$ 0.1 \\
\midrule

\multicolumn{6}{l}{\textbf{Fixed $b=0.25$}} \\
\midrule
-0.025  & 0.25 & 87.2 $\pm$ 0.1 & 49.7 $\pm$ 0.3 & 77.7 $\pm$ 0.0 & 63.4 $\pm$ 0.1 \\
-0.0125 & 0.25 & 87.2 $\pm$ 0.0 & 48.9 $\pm$ 0.2 & 77.3 $\pm$ 0.1 & 62.8 $\pm$ 0.2 \\
0.0125  & 0.25 & 87.5 $\pm$ 0.0 & 49.8 $\pm$ 0.0 & 77.4 $\pm$ 0.1 & 63.2 $\pm$ 0.1 \\
0.025   & 0.25 & 87.7 $\pm$ 0.1 & 50.4 $\pm$ 0.1 & 77.3 $\pm$ 0.1 & 62.5 $\pm$ 0.1 \\
\midrule

\multicolumn{6}{l}{\textbf{Fixed $b=0.5$}} \\
\midrule
-0.05   & 0.5 & 87.6 $\pm$ 0.0 & 49.9 $\pm$ 0.1 & 77.8 $\pm$ 0.0 & 63.2 $\pm$ 0.0 \\
-0.025  & 0.5 & 87.0 $\pm$ 0.0 & 48.4 $\pm$ 0.2 & 77.4 $\pm$ 0.1 & 63.2 $\pm$ 0.2 \\
-0.0125 & 0.5 & 87.8 $\pm$ 0.1 & 50.3 $\pm$ 0.1 & 77.7 $\pm$ 0.1 & 63.4 $\pm$ 0.2 \\
0.0125  & 0.5 & 87.8 $\pm$ 0.0 & 50.6 $\pm$ 0.1 & 77.8 $\pm$ 0.1 & 63.4 $\pm$ 0.2 \\
0.025   & 0.5 & 87.8 $\pm$ 0.1 & 50.6 $\pm$ 0.2 & 77.7 $\pm$ 0.1 & 63.3 $\pm$ 0.1 \\
0.05    & 0.5 & 87.7 $\pm$ 0.2 & 50.7 $\pm$ 0.3 & 77.6 $\pm$ 0.1 & 63.1 $\pm$ 0.1 \\
\bottomrule
\end{tabular}
\end{table*}

The reweighting proposed by \citet{kim2025predict} increases the masking probability of features with high missingness. This is similar to an inverse propensity score correction  \citep{rosenbaum1983central}. Rare but diagnostically relevant features receive an augmented mask with a higher probability and receive prediction signals, ensuring that their contribution to the gradient signal is not washed out by frequent variables \citep{inversepropensity}. This has been proved useful for imputation benchmarks with synthetic missingness \citep{kim2025predict}.

However, shifting to representation learning settings, where the goals involve classification tasks and beyond, the interpretation cannot be translated word for word from imputation contexts. For example, frequently observed variables may provide reliable supervision and broad physiological coverage. They matter for downstream tasks and serve as reference points for reconstructing other features \citep{heilbroner2025self}, so upweighting rare features could trade away signal as much as it recovers it. Neither direction dominates in our ablation, which would fit such an account.

\paragraph{Input Ablation} Table \ref{tab:input_ablation_results} provides information that replacing hourly information of vital signs and vasopressors to daily decreases performance on downstream predictions. Additionally, the table shows insights that including a zero-imputation on vasopressor did not yield a better performance, suggesting the model handles these absent values effectively through the masking mechanism alone. 

\begin{table*}[h]
\centering
\small
\setlength{\tabcolsep}{6pt}
\renewcommand{\arraystretch}{1.05}
\caption{Input ablation study results (mean $\pm$ standard deviation across 5 random seeds).}
\label{tab:input_ablation_results}
\begin{tabular}{lcccc}
\toprule
\multirow{2}{*}{\textbf{Input type}} &
\multicolumn{2}{c}{\textbf{Mortality}} &
\multicolumn{2}{c}{\textbf{Length of Stay}} \\
\cmidrule(lr){2-3}\cmidrule(lr){4-5}
& \textbf{AUROC} & \textbf{AUPRC} & \textbf{AUROC} & \textbf{AUPRC} \\
\midrule
Without 24-hour information & 85.4 $\pm$ 0.1 & 45.7 $\pm$ 0.2 & 75.6 $\pm$ 0.1 & 60.4 $\pm$ 0.1 \\
Zero-filling                & 87.2 $\pm$ 0.1 & 49.3 $\pm$ 0.2 & 77.3 $\pm$ 0.1 & 62.7 $\pm$ 0.1 \\
AID-MAE                     & 87.7 $\pm$ 0.1 & 49.8 $\pm$ 0.1 & 77.6 $\pm$ 0.1 & 63.1 $\pm$ 0.1 \\
\bottomrule
\end{tabular}
\end{table*}

\section{Discussion}

We presented AID-MAE, a dual-masked autoencoder that explicitly combines an intrinsic missingness mask with augmented masking to learn contextual embeddings directly from incomplete EHR tables. This design reduces the need for prior imputation, or unnecessary computation, while enabling capturing temporal-feature dynamics inherent in clinical signals. Empirically, AID-MAE achieves consistent gains over strong baselines on both MIMIC-IV and PhysioNet Challenge 2012 across three clinical prediction tasks. Our results highlight that learning directly from incomplete EHRs data provides a scalable path toward tabular foundation models in healthcare.

\paragraph{Limitations}

Despite these strengths, AID-MAE has limitations. Informative missingness is currently modeled implicitly. Explicitly capturing missing-not-at-random mechanisms remains an important direction for future work, even though the learned representations are encouraged to be invariant to the missingness distribution \citep{du2024remasker}. We also see two complementary future directions: incorporating structured medical or causal knowledge to better guide contextual dependency learning, and extending AID-MAE to multimodal inputs, such as clinical text and medical images.

\acks{
The authors thank Guillaume Obozinski and Nicolas Boumal for their valuable feedback. DR was supported by the European Union’s Horizon Europe research and innovation programme under the Marie Skłodowska-Curie COFUND grant agreement No 101127936 (DeMythif.AI). LAC is supported by the National Institutes of Health (DS-I Africa U54 TW012043, Bridge2AI OT2 OD032701), the National Science Foundation (ITEST 2148451), the Boston–Korea Innovative Research Project (RS-2024-00403047), and the Korea Health Technology R\&D Project (RS-2024-00439677) through the Korea Health Industry Development Institute, funded by the Ministry of Health \& Welfare, Republic of Korea.}

\bibliography{main}

\newpage
\appendix

\section{Additional Result}\label{app:additional}

\subsection{Complete results}  \label{app:complete}

\begin{figure}[t]
\centering
\includegraphics[width=\textwidth]{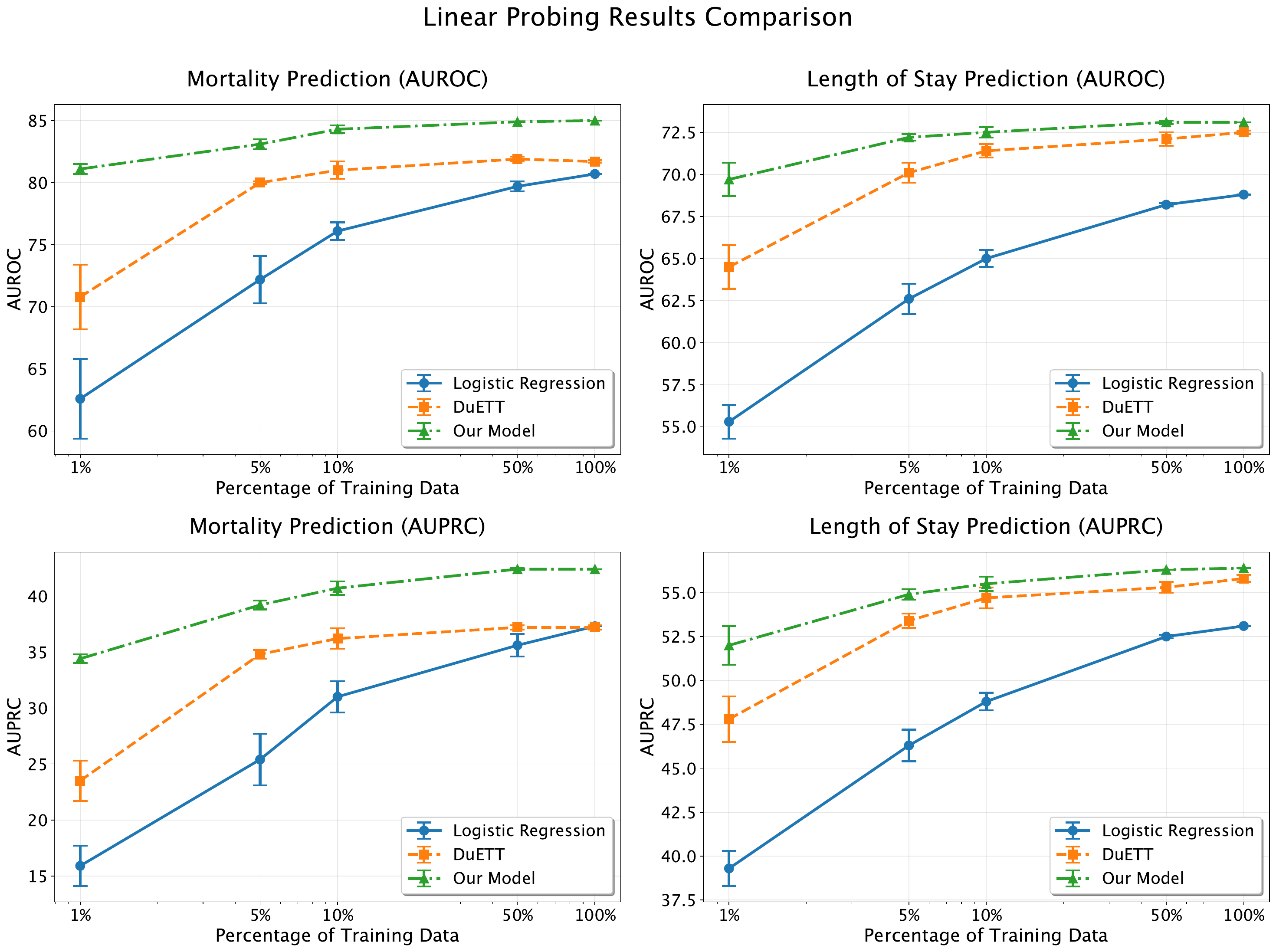}
\caption{Linear probing results for mortality prediction and length of stay prediction tasks. We compare our model against Logistic Regression with median imputation and DuETT across different training data percentages (1\%, 5\%, 10\%, 50\%, 100\%). Results are shown for both AUROC (top row) and AUPRC (bottom row) metrics. Our model consistently outperforms baseline methods across all data regimes and tasks, with particularly strong performance in low-data scenarios. Error bars represent standard deviation across 5 random seeds. The x-axis uses logarithmic scaling to better visualize performance across the range of training data percentages.}
\label{fig:linear_probing_full}
\end{figure}

Tables \ref{tab:mimic_downstream_results} and \ref{tab:physio_downstream_results} show that our model achieves the best performance across both prediction tasks compared to all baselines. Figure \ref{fig:linear_probing_full} further demonstrates that these gains hold consistently for linear probing.

\begin{table*}[t]
\centering
\caption{Downstream task performance on the MIMIC-IV ICU dataset. We compare our proposed model against established baselines across two clinical tasks. Results are reported as mean $\pm$ standard deviation across 5 random seeds. Our model (random masking 25\%) achieves superior performance on both tasks, demonstrating the effectiveness of AID-MAE. Best results are shown in \textbf{bold}. AUROC and AUPRC are reported as percentages with one decimal place.}
\label{tab:mimic_downstream_results}
\small
\begin{tabular}{lcc|cc}
\toprule
& \multicolumn{2}{c}{\textbf{ Mortality}} & \multicolumn{2}{c}{\textbf{Length of Stay}} \\
\cmidrule(lr){2-3} \cmidrule(lr){4-5}
\textbf{Model} & \textbf{AUROC} & \textbf{AUPRC} & \textbf{AUROC} & \textbf{AUPRC} \\
\midrule
Logistic Regression     & 80.7 $\pm$ 0.0 & 37.3 $\pm$ 0.0 & 68.8 $\pm$ 0.0 & 53.1 $\pm$ 0.0 \\
XGBoost                 & 86.7 $\pm$ 0.0 & 47.5 $\pm$ 0.0 & 76.9 $\pm$ 0.0 & 62.2 $\pm$ 0.0 \\
Supervised Transformer  & 83.9 $\pm$ 0.5 & 42.2 $\pm$ 1.3 & 72.3 $\pm$ 0.5 & 56.7 $\pm$ 1.5 \\
\midrule
DuETT                   & 86.4 $\pm$ 0.2 & 47.3 $\pm$ 0.2 & 77.2 $\pm$ 0.0 & 62.8 $\pm$ 0.2 \\
\textbf{AID-MAE (ours)} & \textbf{87.7 $\pm$ 0.1} & \textbf{49.8 $\pm$ 0.1} & \textbf{77.6 $\pm$ 0.1} & \textbf{63.1 $\pm$ 0.1} \\
\bottomrule
\end{tabular}
\end{table*}

\begin{table*}[t]
\centering
\caption{Downstream task performance on the PhysioNet 2012 Challenge dataset. We compare our proposed model against established baselines across two clinical tasks. Results are reported as mean $\pm$ standard deviation across 5 random seeds. Our model (random masking 25\%) achieves superior performance on both tasks, demonstrating the effectiveness of AID-MAE. Best results are shown in \textbf{bold}. AUROC and AUPRC are reported as percentages with one decimal place.}
\label{tab:physio_downstream_results}
\small
\begin{tabular}{lcc|cc}
\toprule
& \multicolumn{2}{c}{\textbf{Mortality}} & \multicolumn{2}{c}{\textbf{Acute Kidney Injury}} \\
\cmidrule(lr){2-3} \cmidrule(lr){4-5}
\textbf{Model} & \textbf{AUROC} & \textbf{AUPRC} & \textbf{AUROC} & \textbf{AUPRC} \\
\midrule
Logistic Regression     & 72.7 $\pm$ 0.0 & 31.3 $\pm$ 0.0 & 74.8 $\pm$ 0.0 & 58.6 $\pm$ 0.0 \\
Supervised Transformer  & 76.7 $\pm$ 1.4 & 34.1 $\pm$ 3.5 & 76.1 $\pm$ 0.9 & 59.6 $\pm$ 1.3 \\
XGBoost                 & 76.9 $\pm$ 0.0 & 36.9 $\pm$ 0.0 & 77.1 $\pm$ 0.0 & 61.8 $\pm$ 0.0 \\
DuETT                   & 77.7 $\pm$ 0.4 & 38.8 $\pm$ 0.9 & 76.7 $\pm$ 0.2 & 61.8 $\pm$ 0.5 \\
\midrule
\textbf{AID-MAE (ours)} & \textbf{78.2 $\pm$ 0.3} & \textbf{39.3 $\pm$ 1.7} & \textbf{77.3 $\pm$ 0.1} & \textbf{62.5 $\pm$ 1.2} \\
\bottomrule
\end{tabular}
\end{table*}

\subsection{Pretraining on Imputed Data} \label{app:mice_result}
Table \ref{tab:mice_ablation} shows that pretraining AID-MAE on MICE-imputed PhysioNet Challenge 2012 data \citep{article,silva2012predicting} gives lower downstream AUROC than pretraining on the incomplete table directly. The masked objective reconstructs values, so imputing first makes the model recover statistical estimates rather than measurements a clinician took.

\begin{table}[t]
\centering
\caption{Pretraining AID-MAE on MICE-imputed data \citep{article}. Architecture, hyperparameters, and splits are held fixed. Mean $\pm$ SD over 5 seeds.}
\small
\begin{tabular}{lcc}
\toprule
\textbf{Setting} & \textbf{Mortality} & \textbf{AKI}  \\
\midrule
With MICE                & 73.9 $\pm$ 1.2 & 75.8 $\pm$ 0.1 \\
\textbf{No Imputation}   & \textbf{78.2 $\pm$ 0.3} & \textbf{77.3 $\pm$ 0.1} \\
\bottomrule
\end{tabular}
\label{tab:mice_ablation}
\end{table}

\subsection{Single-Value Reconstruction Analysis} \label{app:reconstruction}

Imputation in EHR ranges from predicting one missing value, completing entire panels to filling the whole dataset.  
Here we first focus on single-value reconstruction to guide understanding on the pre-training quality.  
Single-value reconstruction tests whether the model can infer one measurement from the remaining context.   
As only one feature is hidden at a time, the task is transparent and easy to interpret.  
This procedure relies purely on the self-supervised training signal of the masked autoencoder, without requiring extra task-specific heads.

We consider each input as a feature vector \(\mathbf{x}\in\mathbb{R}^p\) together with a mask \(\mathbf{m}\in\{0,1\}^p\),  
where \(m_j=1\) means feature \(j\) is observed and \(m_j=0\) means it is missing.  
The mask ensures that the model never uses values that are unavailable.

For evaluation, we hide one additional observed feature at a time.  
Let \(j\) be such a feature with \(m_j=1\).  
We set \(m_j=0\), feed the masked vector into the model, and obtain a reconstruction
\[
\hat{\mathbf{x}} = f(\mathbf{x}\odot\mathbf{m}).
\]
We therefore compute feature-wise evaluation metrics by comparing the predicted value 
\(\hat{x}_j\) with the corresponding ground-truth observation \(x_j\) for each feature.

Our model has consistently outperformed XGBoost \citep{chen2016xgboost}, one of the strongest baselines for single tabular data imputation, across standard error and goodness-of-fit criteria on MIMIC-IV. As shown in Table~\ref{tab:single_val_reconstruction}, our model achieves superior performance in 19 out of 20 most frequently taken clinical features across all metrics, demonstrating consistent advantages in normalized root mean squared error (NRMSE), normalized mean absolute error (NMAE), and coefficient of determination (R\textsuperscript{2}). This advantage is clinically meaningful, as accurate imputation of missing laboratory values is still critical for robust clinical prediction and unbiased model evaluation in many clinical scenarios.

\begin{table*}[h]
\footnotesize
\centering
\begin{tabular}{|l|cc|cc|cc|}
\hline
\textbf{Feature} & \multicolumn{2}{c|}{\textbf{NRMSE}} & \multicolumn{2}{c|}{\textbf{NMAE}} & \multicolumn{2}{c|}{\textbf{R\textsuperscript{2}}} \\
\hline
& \textbf{Model} & \textbf{XGBoost} & \textbf{Model} & \textbf{XGBoost} & \textbf{Model} & \textbf{XGBoost} \\
\hline
Glucose finger stick & \textbf{0.2030} & 0.2033 & \textbf{0.1516} & 0.1523 & \textbf{0.4806} & 0.4787 \\
Potassium (serum) & \textbf{0.1736} & 0.2026 & \textbf{0.1328} & 0.1591 & \textbf{0.5883} & 0.4391 \\
Sodium (serum) & \textbf{0.0676} & 0.0846 & \textbf{0.0416} & 0.0626 & \textbf{0.9362} & 0.9000 \\
Chloride (serum) & \textbf{0.0556} & 0.0702 & \textbf{0.0348} & 0.0514 & \textbf{0.9567} & 0.9310 \\
Hemoglobin & \textbf{0.0571} & 0.0622 & \textbf{0.0391} & 0.0429 & \textbf{0.9602} & 0.9528 \\
Hematocrit (serum) & \textbf{0.0636} & 0.0662 & \textbf{0.0429} & 0.0452 & \textbf{0.9495} & 0.9452 \\
HCO3 (serum) & \textbf{0.0621} & 0.0837 & \textbf{0.0390} & 0.0628 & \textbf{0.9467} & 0.9032 \\
Creatinine (serum) & \textbf{0.0812} & 0.0915 & \textbf{0.0458} & 0.0522 & \textbf{0.9108} & 0.8868 \\
Anion gap & \textbf{0.0859} & 0.1126 & \textbf{0.0544} & 0.0857 & \textbf{0.8982} & 0.8250 \\
BUN & \textbf{0.0862} & 0.0943 & \textbf{0.0550} & 0.0610 & \textbf{0.9077} & 0.8897 \\
Glucose (serum) & \textbf{0.1797} & 0.1860 & \textbf{0.1296} & 0.1355 & \textbf{0.5726} & 0.5419 \\
Magnesium & \textbf{0.1966} & 0.2024 & \textbf{0.1482} & 0.1540 & \textbf{0.4629} & 0.4304 \\
Phosphorous & \textbf{0.1508} & 0.1583 & \textbf{0.1150} & 0.1217 & \textbf{0.6778} & 0.6449 \\
Calcium non-ionized & \textbf{0.1549} & 0.1644 & \textbf{0.1164} & 0.1243 & \textbf{0.6769} & 0.6360 \\
Platelet Count & \textbf{0.1098} & 0.1177 & \textbf{0.0705} & 0.0767 & \textbf{0.8326} & 0.8074 \\
WBC & \textbf{0.1475} & 0.1547 & \textbf{0.1037} & 0.1096 & \textbf{0.7014} & 0.6716 \\
PH (Arterial) & \textbf{0.0588} & 0.0616 & \textbf{0.0316} & 0.0341 & \textbf{0.9526} & 0.9479 \\
Arterial O2 pressure & \textbf{0.1958} & 0.2124 & \textbf{0.1426} & 0.1583 & \textbf{0.4576} & 0.3614 \\
Arterial CO2 Pressure & \textbf{0.0433} & 0.0500 & \textbf{0.0235} & 0.0276 & \textbf{0.9717} & 0.9622 \\
Arterial Base Excess & \textbf{0.0304} & 0.0305 & 0.0224 & \textbf{0.0206} & \textbf{0.9859} & 0.9858 \\
\hline
\end{tabular}
\caption{Performance comparison between our proposed model and XGBoost baseline on the top 20 frequently collected clinical features ranked by frequencies. Metrics include normalized root mean squared error (NRMSE), normalized mean absolute error (NMAE), and coefficient of determination (R\textsuperscript{2}). Normalization is based on the value range. Bold values indicate superior performance (lower NRMSE/NMAE or higher R\textsuperscript{2}). Our proposed model demonstrates consistent superiority across 19 out of 20 features across all metrics.}
\label{tab:single_val_reconstruction}
\end{table*}

\section{Dataset Details}\label{app:data}

\subsection{Dataset Details}

The selected set of 61 features in MIMIC-IV in Table \ref{tab:all_features} provides complementary views on a patient's physiological state and disease progression.

\begin{table*}[!h]
\centering
\scriptsize  
\setlength{\tabcolsep}{3pt}
\renewcommand{\arraystretch}{1.06}
\begingroup
\hyphenpenalty=10000\exhyphenpenalty=10000\pretolerance=10000
\caption{Features Used in the Experiments}
\begin{tabular}{@{}p{0.10\textwidth} >{\raggedright\arraybackslash}p{0.36\textwidth}
                p{0.10\textwidth} >{\raggedright\arraybackslash}p{0.36\textwidth}@{}}
\toprule
\textbf{ID} & \textbf{Feature} & \textbf{ID} & \textbf{Feature} \\
\midrule

\multicolumn{4}{@{}l}{\textbf{Vital Signs \& Physiological Monitoring}} \\
\midrule
220045 & Heart Rate & 220050 & Arterial Blood Pressure systolic \\
220051 & Arterial Blood Pressure diastolic & 220210 & Respiratory Rate \\
220277 & O$_2$ saturation (pulse oximetry) & 223761 (223762) & Temperature ($^\circ$F \& $^\circ$C) \\

\midrule
\multicolumn{4}{@{}l}{\textbf{Blood Gas Panel}} \\
\midrule
220224 & Arterial pO$_2$ & 220227 & Arterial O$_2$ saturation \\
220235 & Arterial pCO$_2$ & 220274 & Venous pH \\
223679 & Venous TCO$_2$ (calc) & 223830 & Arterial pH \\
224828 & Arterial Base Excess & 225698 & Arterial TCO$_2$ (calc) \\
226062 & Venous pCO$_2$ & 226063 & Venous pO$_2$ \\
227073 & Anion Gap & 227443 & Bicarbonate (HCO$_3$) \\
225668 & Lactic Acid & & \\

\midrule
\multicolumn{4}{@{}l}{\textbf{Basic Metabolic Panel (BMP)}} \\
\midrule
220602 & Chloride & 220615 & Creatinine \\
220621 & Glucose (serum) & 220645 & Sodium (serum) \\
225624 & Blood Urea Nitrogen & 225625 & Calcium (serum) \\
225664 & Glucose (finger\mbox{-}stick) & 226534 & Sodium (whole blood) \\
226537 & Glucose (whole blood) & 227442 & Potassium (serum) \\
227464 & Potassium (whole blood) & 220635 & Magnesium \\
225677 & Phosphate & & \\

\midrule
\multicolumn{4}{@{}l}{\textbf{Complete Blood Count (CBC)}} \\
\midrule
220228 & Hemoglobin & 220545 & Hematocrit (serum) \\
226540 & Hematocrit (calc) & 220546 & White Blood Cells \\
227457 & Platelets & & \\

\midrule
\multicolumn{4}{@{}l}{\textbf{CBC with Differential}} \\
\midrule
225639 & Basophils (\%) & 225640 & Eosinophils (\%) \\
225641 & Lymphocytes (\%) & 225642 & Monocytes (\%) \\
225643 & Neutrophils (\%) & & \\

\midrule
\multicolumn{4}{@{}l}{\textbf{Coagulation Panel}} \\
\midrule
227465 & Prothrombin Time & 227466 & Partial Thromboplastin Time \\
227467 & INR & 227468 & Fibrinogen \\

\midrule
\multicolumn{4}{@{}l}{\textbf{Liver Function Test (LFT) Panel}} \\
\midrule
220587 & Aspartate Aminotransferase (AST) & 220644 & Alanine Aminotransferase (ALT) \\
225612 & Alkaline Phosphatase & 225690 & Total Bilirubin \\

\midrule
\multicolumn{4}{@{}l}{\textbf{Cardiac Enzymes Panel}} \\
\midrule
220632 & Lactate Dehydrogenase & 225634 & Creatine Kinase (CK) \\
227445 & CK\mbox{-}MB isoenzyme & 227429 & Troponin\mbox{-}T \\
227456 & Albumin & & \\

\midrule
\multicolumn{4}{@{}l}{\textbf{Vasopressor Medications}} \\
\midrule
221289 & Epinephrine (mcg/min) & 221906 & Norepinephrine (mcg/min) \\
221662 & Dopamine (mcg/min) & 221749 & Phenylephrine (mcg/min) \\
222315 & Vasopressin (units/min) & & \\

\bottomrule
\label{tab:all_features}
\end{tabular}
\endgroup
\end{table*}

For mortality prediction, vital signs, blood gas features, and vasopressor administration directly capture hemodynamic instability and organ dysfunction, while laboratory panels such as the Basic Metabolic Panel (BMP), Complete Blood Count (CBC), coagulation studies, Liver Function Tests (LFT), and cardiac enzyme assays reflect metabolic imbalance, infection response, and multi-organ failure. They are all central indicators of adverse ICU outcomes \citep{Moor2020.08.31.20185207, earlywarning}. For length-of-stay prediction, persistent abnormalities in electrolytes, hematological markers, and liver or kidney function are associated with slower recovery trajectories, while ongoing vasopressor dependence often signals prolonged ICU admission \citep{pmlr-v259-bellamy25a}. 

Additionally, in the broader context of developing clinical foundation models, it is essential to include as many heterogeneous features as possible rather than restricting analyses to a single subset, such as laboratory tests \citep{restrepo2025representationlearninglabvalues}, hence our choice of feature selection. To remain compatible with MIMIC-IV, we retain only those PhysioNet 2012 Challenge features that have a direct counterpart in MIMIC-IV.

\subsection{Preprocessing Details}\label{apd:second}

Clinical time series are recorded at uneven times.  A direct feed would give the model very long
and very sparse sequences. \citet{labach2023duett} show that discretising time series into fixed time bins with the last observed value yields strong performance across clinical prediction tasks. Their ablation study finds that carrying forward the most recent measurement, rather than using an average or interpolation, preserves sharp physiological signals and aligns with clinical intuition. \citet{shukla2020survey} support this finding, highlighting that discretisation with simple carry-forward is both effective and widely used in practice.

Following this, we transform each ICU stay into a sequence of daily rows. Each row summarizes one calendar day of the patient's record. For laboratory values, we retain the last recorded result of each test within that day. This design mirrors standard clinical workflows, where the most recent lab panel is typically used for decision making. Within each row, we also include an hourly representation of the five vital signs, oxygen saturation, and five vasopressors infusion rates. These are recorded at hourly resolution by selecting the last observed value up to that hour. We additionally include a reference value for each lab, which corresponds to the most recent previous corresponding lab result that is recorded prior to the day. This aims to align with real-life clinical setting and provide a longitudinal baseline

Each numeric entry is paired with a time stamp that encodes its recency. We express this as the number of hours before midnight, rounded to one decimal place. For example, a value recorded at 21:00 today is encoded as $3.0$, while a reference lab value taken at 20:30 the day before is encoded as $27.5$. For vasopressors that are carried forward across hours, we assign each event a time stamp corresponding to the midpoint of the hour in which it was forwarded. We first winsorize each numeric feature at the 5th and 95th percentiles to reduce extreme outliers \citep{wilcox2011introduction}, then apply a per-feature min--max normalization. If the values are not recorded, those values will be represented as a missing entry.

We intentionally replace dopamine with norepinephrine equivalent dose since simple arithmetic on the features in the resulted table can recover the dopamine rate. Let $\mathrm{Epi}$, $\mathrm{NE}$, and $\mathrm{Phen}$ be the infusion rates of epinephrine, norepinephrine, and phenylephrine in $\mu\text{g}\,\text{kg}^{-1}\,\text{min}^{-1}$, and let $\mathrm{Dop}$ and $\mathrm{Vas}$ be the rates of dopamine and vasopressin in the same units and in $\text{U}\,\text{min}^{-1}$ respectively. Following standard practice \citep{JENTZER2018416}, we compute

\[
\mathrm{NE}_{\text{eq}} = \mathrm{NE} + \mathrm{Epi} + \frac{\mathrm{Dop}}{150} + \frac{\mathrm{Phen}}{10} + 2.5\,\mathrm{Vas}.
\]
Lastly, rows that do not contain any laboratory measurements are discarded, as they convey thin physiological signal \citep{10.1001/jamanetworkopen.2018.4521}. This filter removes approximately seven percent of input samples in MIMIC-IV and prevents the model from over-fitting patterns driven purely by sparsity.

We split our curated dataset into training set and test set based on their time stamps. Admission prior to the year 2179 constitute the training set, and after as the test set. We note that year 2179 is a de-identified number by MIMIC-IV \citep{johnson_mimic-iv_2023}, effectively yielding a reproducible random split. We further split 20\% of the training set as the validation set. Disjoint subject\_ids are ensured across the train, validation and test set.

\section{Architectural Details}  \label{app:architecture}

This section presents the comprehensive architectural and training specifications for our masked autoencoder framework, covering pretraining, linear probing, and fine-tuning.

\subsection{Pretraining Architecture and Configuration}

\paragraph{Pretraining Architecture}

The masked autoencoder follows a Vision Transformer-inspired encoder-decoder architecture adapted for tabular data. The encoder consists of a configurable number of transformer blocks with the following specifications:

\begin{itemize}
    \item \textbf{Embedding dimension}: $d_{embed} = 64$ 
    \item \textbf{Encoder depth}: $L_{enc} = 8$ transformer blocks 
    \item \textbf{Number of attention heads}: $h = 8$ (proportional to embedding dimension)
    \item \textbf{MLP ratio}: $r_{mlp} = 4.0$ (hidden dimension = $4 \times d_{embed}$)
    \item \textbf{Decoder embedding dimension}: $d_{dec} = 64$ (matches encoder)
    \item \textbf{Decoder depth}: $L_{dec} = 4$ transformer blocks
    \item \textbf{Decoder attention heads}: $h_{dec} = 4$
\end{itemize}

The model employs a specialized CombinedEmbed module that processes value-time pairs by projecting each component to the full embedding dimension and combining them additively, effectively reducing the sequence length by half while preserving temporal information.

\paragraph{Training configuration} The pretraining uses AdamW \citep{loshchilov2019decoupledweightdecayregularization} optimizer with a base learning rate of $lr_{base} = 1 \times 10^{-3}$ and weight decay of $\lambda = 0.05$. The learning rate follows a cosine annealing schedule \citep{loshchilov2016sgdr} with 20-40 warmup epochs, decaying to a minimum learning rate of $lr_{min} = 1 \times 10^{-5}$. \\

\textbf{Training Dynamics}: 
\begin{itemize}
    \item \textbf{Batch size}: $B = 64$ samples per batch with gradient accumulation support
    \item \textbf{Maximum epochs}: $E_{max} = 400 $
    \item \textbf{Loss function}: Mean squared error
\end{itemize}

\subsection{Linear Probing Methodology}

Linear probing evaluation extracts frozen representations from the pretrained encoder to assess learned feature quality. The methodology involves:

\textbf{Feature Extraction}: CLS token embeddings ($d_{embed} = 64$-dimensional) are extracted from the pretrained encoder without any fine-tuning of the encoder parameters.

\textbf{Classifier Configuration}: Scikit-learn's LogisticRegression with liblinear solver, L2 regularization ($C = 1.0$).

\textbf{Evaluation Protocol}: 
\begin{itemize}
    \item \textbf{Data fractions}: We use \(f \in \{1\%, 5\%, 10\%, 50\%, 100\%\}\) of the available training data.
    \item \textbf{Seeds}: 5 independent runs (2020-2024) for statistical robustness
    \item \textbf{Metrics}: AUROC and AUPRC for binary classification tasks
    \item \textbf{Tasks}: Mortality and 72-hour length-of-stay prediction for MIMIC-IV, In-hospital mortality and Acute Kidney Injury for PhysioNet 2012 Challenge.
\end{itemize}

\subsection{Fine-tuning Architecture and Training}

The fine-tuning approach utilizes a task-specific classification head appended to the pretrained encoder:

\textbf{Classification Head}: A configurable feedforward network with the following default configuration:
\begin{itemize}
    \item \textbf{Input dimension}: $d_{in} = 64$ (matching encoder embedding dimension)
    \item \textbf{Hidden layers}: 2-layer MLP architecture
    \item \textbf{Hidden dimension}: $d_{hidden} = 32$ neurons per hidden layer
    \item \textbf{Dropout rate}: $p_{drop} = 0.1$
    \item \textbf{Output dimension}: $d_{out} = 1$ (binary classification with sigmoid activation)
\end{itemize}

\paragraph{Fine-tuning Training Configuration}

\begin{itemize}
    \item \textbf{Encoder learning rate}: $lr_{enc} = 1 \times 10^{-5}$ 
    \item \textbf{Classification head learning rate}: $lr_{cls} = 1 \times 10^{-3}$ 
\end{itemize}

\textbf{Optimizer and Regularization}:
\begin{itemize}
    \item \textbf{Optimizer}: AdamW with weight decay of $\lambda = 1 \times 10^{-5}$
    \item \textbf{Batch size}: $B = 128$ samples per batch
    \item \textbf{Maximum epochs}: $E_{max} = 100$ with early stopping
\end{itemize}

\textbf{Training Protocol}:
\begin{itemize}
    \item \textbf{Early stopping}: Patience of 10 epochs based on validation AUROC
    \item \textbf{Validation strategy}: Stratified train/validation/test splits of 64\%, 16\% and 20\%
\end{itemize}

The training framework allows systematic evaluation of self-supervised pretraining effectiveness across different adaptation strategies, from linear probing (with no parameter updates) to full fine-tuning (end-to-end optimization).

\section{Experiment Details}

\subsection{Baseline Details} \label{app:baseline}

\subsubsection{Logistic regression} 
As a linear baseline, we trained logistic regression models with both $\ell_{1}$ and 
$\ell_{2}$ penalties. Prior to training, we applied median imputation to all continuous 
variables to address missing values, ensuring that imputation was performed once on the 
training set and applied consistently to the test set. Model selection was carried out using a grid search over the regularization strength 
$C \in \{0.1, 1.0, 10.0\}$ and penalty type $ \{l_1, l_2\}$. Optimization used the \texttt{liblinear} solver, 
with a maximum of 200 iterations and a convergence tolerance of $10^{-3}$. 
To reduce the impact of random variation, we repeated experiments with five random seeds 
(2020--2024) and report the mean and standard deviation of AUROC and AUPRC on the test set. 

\subsubsection{XGBoost} 

We additionally benchmarked against XGBoost~\citep{chen2016xgboost}, a widely used gradient 
boosted decision tree method that has been shown to perform competitively on tabular 
and clinical time series data. 
The model was tuned with a grid search over 
$\{ \text{learning rate} \in \{0.05, 0.1, 0.2\}, \; \text{max depth} \in \{4, 5, 6\}, \; 
\text{number of estimators} \in \{500, 1000\} \}$, and hyperparameters were selected via 
validation AUROC. For evaluation, the best 
model was applied to the test set to compute AUROC and AUPRC. 

\subsubsection{MICE Imputation} \label{app:mice}
We used scikit-learn's IterativeImputer with BayesianRidge as the default estimator, 10 imputation rounds, and a fixed random seed. Imputation was fit on the training set and applied to validation and test sets. The imputed data was then used to train AID-MAE with identical architecture and hyperparameters, isolating the effect of imputation on the training signal.

\subsubsection{DuETT: Dual Event Time Transformer}

DuETT~\citep{labach2023duett} extends the Transformer architecture to clinical time series by 
explicitly modelling three fundamental dimensions of electronic health record (EHR) data. 
First, temporal dependencies are captured by attention over discretized time bins, 
enabling the model to learn disease trajectories despite irregular sampling. 
Second, event-type relationships are captured by attention across heterogeneous 
clinical variables, allowing the model to integrate information from diverse measurements 
such as vitals and laboratory values. 
Third, DuETT leverages the presence or absence of observations as a predictive signal, 
by jointly reconstructing masked event values and missingness indicators during 
self-supervised pre-training. This design exploits the fact that not measuring a variable 
in itself conveys clinical intent. 

By alternating attention across time and event axes, and by integrating missingness as a 
training signal, DuETT learns robust patient representations that outperform 
state-of-the-art baselines in both full fine-tuning and limited-label regimes. One limit that we currently address is that DuETT does not directly attend different features across time, losing information on temporal interactions among features.

We follow the training protocol of DuETT, which consists of a 
self-supervised pre-training phase followed by supervised adaptation. 

\paragraph{Pre-training} Following~\citep{labach2023duett}, we conducted self-supervised pre-training for 
300~epochs using the AdamW optimizer. The learning rate was scheduled with linear warmup 
followed by inverse square-root decay, and gradient clipping was applied at 1.0 to stabilize 
training. At each iteration, one event type and one time bin were masked and the model was 
trained to reconstruct both event values and their presence. Inputs were normalized to zero mean and unit variance, with 
outliers clipped at three median absolute deviations from the median, and time-bin aggregation 
used the last observed value. Dropout was applied in attention and feedforward layers as in the 
original implementation.

\paragraph{Fine-tuning} 
All encoder and classification head parameters were updated jointly using a single AdamW optimizer, 
without differential learning rates. This choice follows the official DuETT implementation. In practice, this design is 
stabilized by the combination of linear warmup and inverse square-root decay, rather than by 
assigning separate learning rates to encoder and head. This is feasible since DuETT's 
representation token is of comparatively high dimension (approximately 1.5k), allowing sufficient 
capacity in the head to adapt during supervised training. We fine-tuned for the same number of 
epochs as reported in~\citep{labach2023duett} (30 epochs on MIMIC-IV data), 
ensuring comparable training budgets. To further enhance robustness, we adopted the procedure of 
the paper, averaging the weights of the five best-performing checkpoints (ranked by 
validation AUROC) to obtain the final model. This weight-averaging strategy was shown to improve 
stability over single-checkpoint selection. Importantly, to rule out undertraining as a confound, 
we additionally performed a systematic learning rate sweep over 
$\{10^{-3}, 10^{-4}, 10^{-5}, 10^{-6}\}$ for both fine-tuning and linear probing, and report the 
best results across seeds. 

\paragraph{Linear probing} 
To directly assess representation quality, we froze the encoder and trained only a linear 
classification head (a single fully connected layer without hidden units). This corresponds to 
logistic regression applied to the fixed patient representation produced by the DuETT encoder. 
Since the encoder output dimension in DuETT is large, linear probing provides a stringent test of 
the quality of pre-trained representations without relying on capacity in the task head.

\end{document}